\documentclass{article} 
\usepackage[final]{colm2026_conference}

\usepackage{microtype}
\usepackage{hyperref}
\usepackage{url}
\usepackage{booktabs}
\usepackage{lineno}

\usepackage{graphicx}
\usepackage{xcolor}
\usepackage{amsthm}

\usepackage{amsmath}

\usepackage{comment}
\usepackage[skip=3pt]{caption}
\usepackage{wrapfig}
\usepackage{titlesec}
\titlespacing{\section}{0pt}{4pt}{2pt}
\titlespacing{\subsection}{0pt}{2pt}{1pt}
\usepackage[most]{tcolorbox}
\usepackage{natbib}
\usepackage{graphicx}
\usepackage{multirow}
\usepackage{array}

\definecolor{darkblue}{rgb}{0, 0, 0.5}
\hypersetup{colorlinks=true, citecolor=darkblue, linkcolor=darkblue, urlcolor=darkblue}

\definecolor{darkblue}{rgb}{0, 0, 0.5}
\hypersetup{colorlinks=true, citecolor=darkblue, linkcolor=darkblue, urlcolor=darkblue}

\title{A Statistical Framework for Auditing Behavioral Dependence and Induced Bias in LLM Judges}

\author{
\bfseries
Chenchen Kuai\textsuperscript{1}\thanks{Equal contribution.},\ 
Jiwan Jiang\textsuperscript{1}\footnotemark[1],\ 
Zihao Zhu\textsuperscript{1},\ 
Hao Wang\textsuperscript{1},\ 
Keshu Wu\textsuperscript{1},\ 
Zihao Li\textsuperscript{2},\\
\bfseries
Yunlong Zhang\textsuperscript{1},\ 
Chenxi Liu\textsuperscript{3},\ 
Zhengzhong Tu\textsuperscript{1},\ 
Zhiwen Fan\textsuperscript{1},\ 
\& Yang Zhou\textsuperscript{1}\thanks{
Corresponding author: \texttt{yangzhou295@tamu.edu}
}\\[0.4em]
\normalfont
\textsuperscript{1}Texas A\&M University,\ 
\textsuperscript{2}Marquette University,\ 
\textsuperscript{3}University of Utah
}

\begin{document}

\ifcolmsubmission
\linenumbers
\fi

\maketitle
\lhead{Published as a conference paper at COLM 2026}

\begin{abstract}
The rapid growth of the large language model (LLM) ecosystem raises a critical question: are seemingly diverse models truly independent? Shared pretraining data, distillation, and alignment pipelines can induce hidden behavioral dependencies, latent entanglement, that undermine multi-model systems such as LLM-as-a-judge pipelines and ensemble verification, which implicitly assume independent signals. In practice, this manifests as correlated reasoning patterns and synchronized failures, where apparent agreement reflects shared error modes rather than independent validation. To address this, we develop a statistical framework for auditing behavioral entanglement among black-box LLMs. Our approach introduces a multi-resolution hierarchy that characterizes the joint failure manifold through two information-theoretic metrics: (i) a Difficulty-Weighted Behavioral Entanglement Index (BEI), which amplifies synchronized failures on easy tasks, and (ii) a Cumulative Information Gain (CIG) metric, which captures directional alignment in erroneous responses. Through experiments on 18 LLMs from 6 model families, we identify statistically significant behavioral entanglement. Such behavioral dependence is associated with judge over-endorsement bias on a disjoint MMLU-Pro evaluation set ($\rho=0.508$ for BEI and $0.520$ for CIG; $p<0.01$).
The association further transfers to the MATH-500 benchmark
($\rho=0.441$ and $0.457$; $p<0.05$), providing cross-benchmark evidence
that the identified dependency structure generalizes beyond the data and
response format used for its estimation. Finally, we demonstrate a practical use case of entanglement through
de-entangled verifier ensemble reweighting, achieving 3.5 and 2.6
percentage-point gains in accuracy and precision, respectively, over
majority voting.
\end{abstract}

\section{Introduction}

The rapid expansion of Large Language Models (LLMs) has outpaced our ability to verify their structural independence. This is increasingly critical as modern AI systems rely on multi-model infrastructures such as LLM-as-a-judge pipelines \citep{gu2026survey}, ensemble evaluation frameworks, and redundancy-based safety mechanisms. For example, in LLM-as-a-judge pipelines, multiple models may unanimously endorse an incorrect answer, where agreement is often treated as evidence of correctness. However, if models share training lineage or alignment signals, such agreement may instead reflect correlated failure rather than independent verification. As a result, these systems implicitly assume independent signals, while the apparent diversity of LLMs may be misleading.

Many LLMs share substantial portions of their training pipelines, overlapping pre-training corpora, shared alignment procedures \citep{zhou2023don}, and widespread knowledge distillation \citep{cho2019efficacy, park2019relational}, which can induce hidden behavioral dependencies. We term this phenomenon \textbf{latent entanglement}, whereby ostensibly independent models exhibit synchronized reasoning patterns and similar failure modes despite being developed separately \citep{ye2025justice}. This poses a fundamental risk for multi-LLM systems: apparent agreement may reflect a \emph{consensus of correlated errors} rather than independent verification, leading to overestimated reliability and shared blind spots \citep{zheng2023judging, liu2024llms, liu2024calibrating, chen2025beyond,zhang2026grounding}. Existing analyses typically rely on descriptive behavioral signals, such as output overlap, embedding similarity, or agreement-based scores~\citep{liu2024aligning, rosas2019quantifying}; however, these measures are not identifiable with respect to independence, as similar outputs may arise from either independent reasoning or latent coupling induced by shared data, benchmark contamination, or evaluator self-preference across related model families~\citep{xu2024benchmark}. Moreover, because such signals operate at the level of observable outputs or explanations, they can be strategically perturbed without altering the underlying decision process, thereby masking dependence.

To address this limitation, we adopt a statistical perspective that explicitly tests whether observed agreement is consistent with independent reasoning, leveraging principled null models and uncertainty quantification. Our key observation is that while correct answers may naturally converge, the structure of errors is more informative: under independence, failures should disperse across a large hypothesis space, whereas coincident failures, especially on easy tasks or along identical incorrect options, provide statistically significant evidence of latent dependence. We formalize this intuition by analyzing the \textbf{joint failure manifold} through a hierarchy of behavioral synchronization.

Specifically, this paper makes the following contributions: (1) \textbf{A Multi-Resolution Statistical Framework for Behavioral Entanglement.} We propose a hierarchical statistical framework for auditing dependencies within the LLM failure manifold. This hierarchy integrates two novel information-theoretic metrics: the \textit{Difficulty-Weighted Behavioral Entanglement Index} for binary failure synchronization, with more emphasis on the joint failures on easy questions and \textit{Cumulative Information Gain} for directional error alignment, providing a robust toolset to detect shared behavior entanglement.(2) We conduct extensive experiments on 18 LLMs from six model families to identify behavioral entanglement and its impact on LLM-as-a-judge evaluation. By leveraging model outputs and cross-model verification signals, we statistically characterize entanglement patterns and demonstrate their association with drop in judge precision. (3) We demonstrate a practical use case for \textbf{Verifier Ensembles through De-entangled Ensemble Reweighting}. We adopt a mitigation algorithm that utilizes entanglement scores to regularize ensemble fusion. By  reweighting verifiers based on their statistical independence from the target model, our method significantly reduces bias and restores the accuracy of multi-model verification.

\section{Related Work}
This study connects three related topics: shared model lineage, bias in LLM-as-a-judge systems, and behavioral signals used to infer dependence among models. Together, they motivate our focus on whether agreement among black-box LLMs is trustworthy when dependencies are not accounted for.

\subsection{Knowledge Distillation and Shared Model Lineage}
A primary source of latent dependence is knowledge distillation and synthetic supervision \citep{wang2025retrieval}. Student models can retain detectable traces of their teachers even when the teacher is accessible only as a black box. For example, \citet{wadhwa2025taught} show that student outputs often preserve identifiable teacher footprints. This concern is reinforced by the widespread use of model-generated instruction data in alignment pipelines~\citep{dubois2023alpacafarm,ji2023vicunaner,peng2023instruction}. At the ecosystem level, recursive reuse of synthetic data can further homogenize future models~\citep{shumailov2024ai}. Together, these trends suggest that behavioral agreement across models may reflect inherited decision structure rather than independent reasoning.

\subsection{LLM-as-a-Judge and Evaluation Biases}
This issue is especially important in the LLM-as-a-judge setting, where strong models are used to evaluate others~\citep{zheng2023judging,wang2024pandalm, kuai2026cyportqa}. Recent work shows that such judges are not neutral. \citet{wataoka2024self} identify self-preference bias, whereby judges favor outputs that are more familiar to them, while \cite{chen2025beyond} show that judges can systematically prefer outputs from related models, including inherited students or models from the same family. As a result, agreement between a judge and a target model cannot automatically be interpreted as independent verification~\citep{li2026preference}. Our work is motivated by precisely this failure mode.

\subsection{Contamination and the Limits of Output-Level Signals}
A related line of work examines benchmark contamination and other output-level signals that may confound model comparison~\citep{deng2024investigating, dong2024generalization, dekoninck2024constat}. Contamination arises when benchmark examples, or close variants of them, appear in training corpora, thereby inflating apparent model capability and agreement~\citep{sainz2023nlp,magar2022data,xu2024benchmark}. Existing diagnostics, such as overlap- or perplexity-based heuristics, can detect some forms of direct leakage~\citep{shi2024detecting}, but they remain limited for paraphrased contamination and do not determine whether multiple models are behaviorally independent. More broadly, observable agreement or similarity alone is not identifiable with respect to independence: models may agree because they reason independently, because they share contaminated benchmark exposure, or because they inherit common decision structure from shared training pipelines. This limitation motivates the need for a statistical framework that moves beyond surface-level signals and directly tests for dependence through the structure of model errors.



\begin{figure}
    \centering\includegraphics[width=1\linewidth,height=0.6\textheight,keepaspectratio]{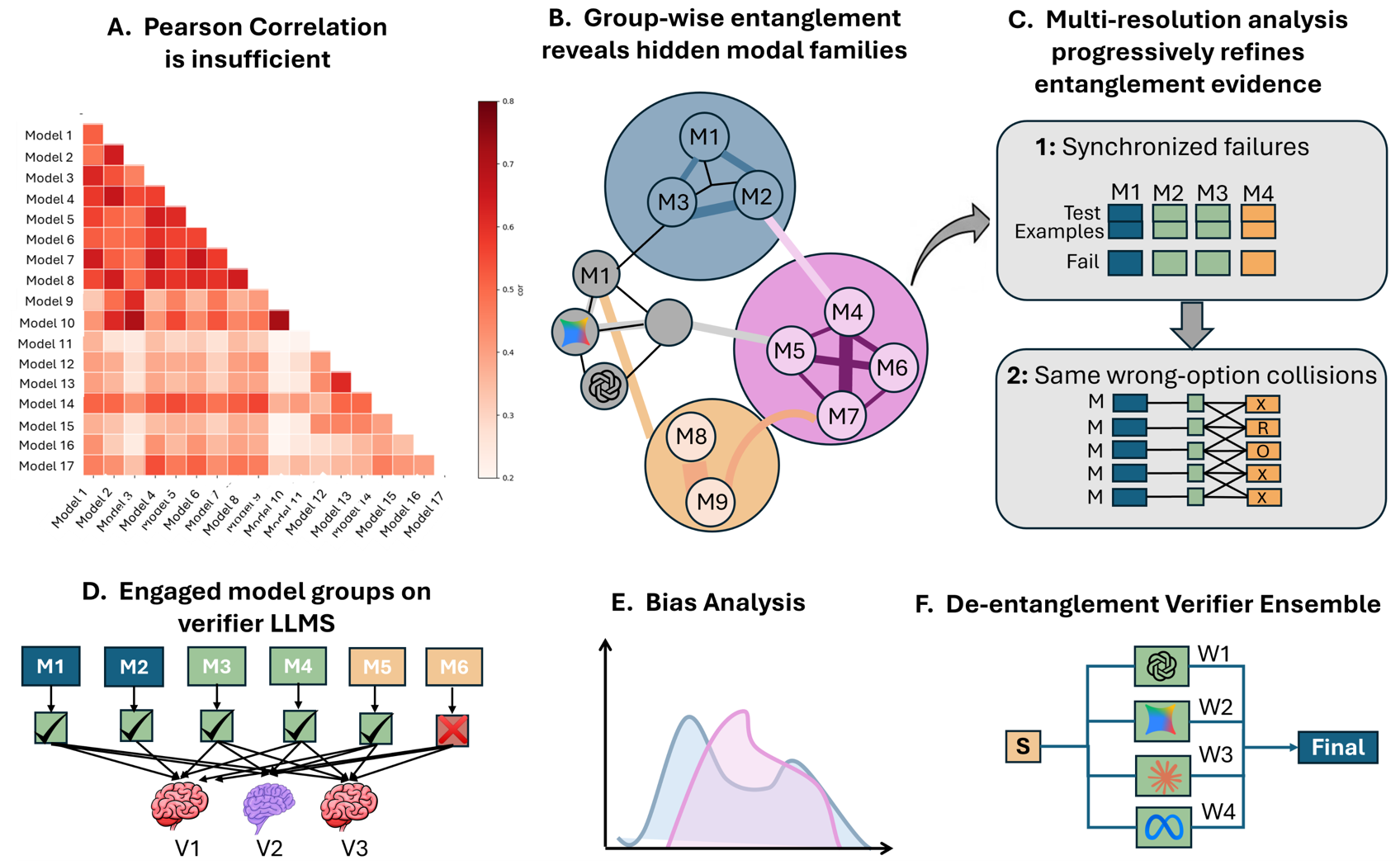}
    \caption{A Multi-Resolution Entanglement Hierarchy for Behavioral Dependence}
    \label{fig:Entanglement Hierarchy}
\end{figure}

\section{Statistical Framework for Auditing Behavioral Entanglement}

Our goal is to determine whether a group of large language models behaves as statistically independent reasoning agents or exhibits hidden structural dependence due to shared training data, distillation, or architectural lineage. Given $M$ black-box models evaluated on $N$ tasks, we test whether their outputs are consistent with independent error processes or display higher-order synchronization indicative of latent entanglement.

A key difficulty is that agreement does not imply dependence: correct answers naturally converge because tasks typically have a single solution. In contrast, \textit{errors} occupy a much larger hypothesis space. When multiple models fail in the same way across many tasks, especially easy tasks, the probability of such synchronization under independence becomes extremely small. This motivates focusing on the \textbf{failure manifold}, where dependence signals become statistically identifiable.

The following subsections describe the two levels of the hierarchy
(see Figure~1) and its downstream application. Level 1 introduces the
Difficulty-Weighted Behavioral Entanglement Index (BEI$_w$) for synchronized
failures, while Level 2 introduces Cumulative Information Gain (CIG) for
directional error alignment. We then examine the downstream association
with judge bias and develop an entanglement-aware verifier reweighting
strategy.

\subsection{Level 1: Failure Synchronization and Pairwise Behavioral Entanglement}

We begin with a coarse manifestation of behavioral dependence: the excess occurrence of co-failures across tasks. Even under behavioral independence, two models are more likely to fail together on intrinsically difficult tasks, simply because both models have higher marginal failure probabilities on such tasks. Therefore, whether an observed co-failure is excessive cannot be assessed without accounting for task difficulty. We address this by defining behavioral independence conditionally on task difficulty. Deviations from this conditional-independence null provide evidence of pairwise behavioral entanglement and are subsequently quantified by the proposed BEI metric.

The assessment is conducted on tasks with objectively verifiable answers. Consider $M$ models evaluated across $T$ tasks. For each model $m$ and task $t$, let $Y_{m,t}\in\{0,1\}$ denote the binary error indicator, where $Y_{m,t}=1$ indicates that model $m$ answers task $t$ incorrectly. For a focal model pair $(i,j)$, we characterize the empirical difficulty of task $t$ using the failure rate among the remaining models:
\begin{equation}
d_t
=
\frac{1}{M-2}
\sum_{m\notin\{i,j\}} Y_{m,t}.
\label{eq:difficulty}
\end{equation}
Excluding the focal pair prevents the two outcomes being tested from mechanically entering their own conditioning variable. The corresponding task easiness is denoted as $a_t = 1-d_t$, so that larger $a_t$ corresponds to an easier task.

\noindent
\textbf{Conditional Independence Null.}
We define the null hypothesis as behavioral independence conditional on task difficulty. Intuitively, among tasks of the same difficulty, the correctness outcomes of two models are assumed to be independent: once task difficulty is known, observing whether one model succeeds or fails provides no additional information about the other model's outcome. Formally,
\begin{equation}
H_0:\quad
Y_{i,t}\perp\!\!\!\perp Y_{j,t}\mid d_t .
\label{eq:conditional_independence}
\end{equation}

For an individual model $m$, its marginal failure probability on a task of difficulty $d_t$ can be written as $\mathbb{P}(Y_{m,t}=1\mid d_t)$.
The conditional-independence null therefore implies that the probability that models $i$ and $j$ fail the same task jointly is determined entirely by their respective marginal failure probabilities at that difficulty:
\begin{equation}
\mathbb{P}
\left(
Y_{i,t}=1,\,
Y_{j,t}=1
\mid d_t
\right)
=
\mathbb{P}(Y_{i,t}=1\mid d_t)
\,
\mathbb{P}(Y_{j,t}=1\mid d_t).
\label{eq:conditional_independence_prob}
\end{equation}
To model how the marginal failure probability of each model varies with task difficulty, define its difficulty-response function as
$
p_m(d)
:=
\mathbb{P}(Y_{m,t}=1\mid d_t=d).
\label{eq:difficulty_response}
$
In practice, $p_m(d)$ can be estimated using logistic regression. This formulation is analogous to difficulty calibration in item response theory \citep{bock2021item}: the objective is not necessarily to construct a perfect predictive model, but to account for the systematic variation in marginal failure probability associated with task difficulty through calibration. Considering the co-failure for model pair $(i,j)$ on task $t$ is $Y_{i,t}Y_{j,t}$, which equals one if and only if both models fail task $t$. Under $H_0$, the expected co-failure probability follows directly from conditional independence 
$
\mathbb{E}_0[Y_{i,t}Y_{j,t}\mid d_t] =
p_i(d_t)\,p_j(d_t).
$

\noindent
\textbf{Behavioral Entanglement Index.}
The deviation from this null expectation provides a task-level measure of synchronized failure. A direct aggregation across all evaluated tasks yields the mean co-failure residual,
\begin{equation}
\overline{r}_{ij}
=
\frac{1}{T}
\sum_{t=1}^{T}
\left(
Y_{i,t}Y_{j,t}
-
p_{i,t}p_{j,t}
\right).
\label{eq:uw_bei}
\end{equation}
This quantity measures the average excess co-failure exhibited by a model pair beyond that expected from their individual difficulty-conditioned failure probabilities.

Although the unweighted residual is naturally centered at zero under $H_0$, residuals from different tasks are not equally informative about shared failure modes. On difficult tasks, the expected co-failure probability $p_{i,t}p_{j,t}$ is already high, so the absence of a co-failure can produce a large negative residual. Such negative contributions are difficult to interpret as evidence against behavioral entanglement, since success on a difficult task may reflect sufficient capability rather than genuinely distinct behavior. Conversely, co-failures on easy tasks are more diagnostically informative. When both models are individually expected to succeed, their simultaneous failure is more indicative of a shared blind spot beyond what task difficulty alone would predict. We therefore weight task-level residuals by task easiness, attenuating contributions from difficult tasks while emphasizing unexpected co-failures on easier ones. Let $a_t$ denote task easiness and define the weighting function $a_t^w=(1-d_t)^w$ for $w\geq0$. The proposed $\mathrm{BEI}_w$ is therefore:
\begin{equation}
\mathrm{BEI}_w(i,j)
=
\frac{1}{T}
\sum_{t=1}^{T}
a_t^w
\left(
Y_{i,t}Y_{j,t}
-
p_{i,t}p_{j,t}
\right)
\label{eq:bei}
\end{equation}
The parameter $w$ controls the strength of the easiness emphasis. Setting $w=0$ recovers mean co-failure residual in Eq.~\eqref{eq:uw_bei}, whereas increasing $w$ progressively attenuates contributions from difficult tasks.

\noindent 
\textbf{Proposition 1 (Null centering under easiness weighting).}
Under $H_0$, the task-level co-failure residual is centered at zero. Since the easiness weight is a deterministic function of task difficulty, weighting preserves this zero expectation, and hence
$\mathbb{E}_0[\mathrm{BEI}_w(i,j)]=0$.
A proof is provided in Appendix A.

\noindent
\textbf{Statistical Inference.}
To determine whether the dependence captured by BEI is statistically significant, we test whether the observed excess co-failure is significantly greater than zero. As established above, $\mathrm{BEI}_w(i,j)$ is centered at zero under the conditional-independence null; accordingly, we consider the one-sided alternative $H_1:\mathbb{E}[\mathrm{BEI}_w(i,j)]>0$.

For each task $t$, the fitted difficulty-response functions provide the marginal failure probabilities
$\hat p_{i,t}=\hat p_i(d_t)$ and $\hat p_{j,t}=\hat p_j(d_t)$. we generate a synthetic realization under $H_0$ by independently sampling the two model outcomes from Bernoulli distributions with these probabilities while keeping the observed task difficulties fixed. We then recompute BEI on the simulated outcomes. Repeating this procedure produces the Monte Carlo null distribution of BEI under conditional independence, from which a one-sided $p$-value is obtained by comparing the observed BEI with the simulated values. Since the audit tests multiple model pairs, we further apply the Benjamini--Hochberg procedure to control the false discovery rate (FDR) and report the corresponding adjusted $q$-values \citep{benjamini1995controlling}.

\subsection{Level 2: Directional Failure Collision and Cumulative Information Gain}

Level~1 identifies whether two models fail together more often than expected, but does not distinguish whether their failures follow the same direction. In multiple-choice question (MCQ) settings, two models may both fail while selecting different distractors, whereas repeatedly selecting the same distractor provides more specific evidence of aligned erroneous behavior. However, such collisions can also arise because some distractors are intrinsically more attractive than others. We therefore assess directional alignment relative to a task-specific distractor distribution and quantify excess alignment CIG.

For an MCQ task $t$ with $K-1$ distractors, let
$
S_{m,t}\in\{1,\ldots,K-1\}
$
denote the distractor selected by model $m$ when $Y_{m,t}=1$. For a focal pair $(i,j)$, the attractiveness of distractor $k$ is estimated from the failing responses of the remaining models \citep{chen1999empirical}:
\begin{equation}
\pi_{k,t}
=
\frac{
n_{k,t}^{(-ij)}+\alpha
}{
n_t^{(-ij)}+\alpha(K-1)
},
\qquad
n_{k,t}^{(-ij)}
=
\sum_{\substack{m\notin\{i,j\}\\Y_{m,t}=1}}
\mathbb{I}(S_{m,t}=k),
\label{eq:distractor_probability}
\end{equation}
where $n_t^{(-ij)}=\sum_{m\notin\{i,j\}}Y_{m,t}$ and $\alpha=1$ corresponds to add-one smoothing. Pair exclusion prevents the tested responses from entering their own directional baseline, while smoothing avoids zero probability for distractors not selected by the remaining models.

\noindent
\textbf{Directional Independence Null.}
Conditional on both models failing task $t$, the null assumes that their distractor choices are independent draws from the task-specific distribution:
\begin{equation}
H_0^{\mathrm{dir}}:\quad
\mathbb{P}
\left(
S_{i,t}=k,\,
S_{j,t}=\ell
\mid
Y_{i,t}=Y_{j,t}=1
\right)
=
\pi_{k,t}\pi_{\ell,t}.
\label{eq:directional_independence}
\end{equation}
Define the directional collision indicator
$
Z_{ij,t}^{\mathrm{dir}}
=
\mathbb{I}(S_{i,t}=S_{j,t}).
$
Under $H_0^{\mathrm{dir}}$, the expected collision probability is
\begin{equation}
c_t
=
\mathbb{E}_0
\left[
Z_{ij,t}^{\mathrm{dir}}
\mid
Y_{i,t}=Y_{j,t}=1
\right]
=
\sum_{k=1}^{K-1}\pi_{k,t}^{\,2}.
\label{eq:null_collision_probability}
\end{equation}

\noindent
\textbf{Cumulative Information Gain.}
The deviation
$
Z_{ij,t}^{\mathrm{dir}}-c_t
$
measures excess directional agreement beyond that expected from distractor attractiveness. These deviations are not equally informative: when $c_t$ is large, selecting the same distractor is already likely under the null, whereas a collision with small $c_t$ is more surprising. We therefore weight each residual by its null surprisal, $-\log c_t$.

Let
$
\mathcal{T}_{ij}^{\mathrm{cf}}
=
\{t:Y_{i,t}=Y_{j,t}=1\}
$
denote the set of co-failure tasks. The proposed \textbf{CIG} is
\begin{equation}
\mathrm{CIG}(i,j)
=
\frac{1}{T}
\sum_{t\in\mathcal{T}_{ij}^{\mathrm{cf}}}
\left(-\log c_t\right)
\left(
Z_{ij,t}^{\mathrm{dir}}-c_t
\right).
\label{eq:cig}
\end{equation}
Positive CIG values indicate excess same-distractor alignment, with greater contribution from collisions that are unlikely under directional independence.

\noindent
\textbf{Statistical Inference.}
To determine whether the directional dependence captured by CIG is statistically significant, we test whether the observed excess directional agreement is significantly greater than zero. Accordingly, we consider the one-sided alternative
$
H_1^{\mathrm{dir}}:
\mathbb{E}[\mathrm{CIG}(i,j)]>0.
$ For each co-failure task $t$, we generate a null directional collision as
$
Z_{ij,t}^{\mathrm{dir},(b)}
\sim
\operatorname{Bernoulli}(c_t),
$
and recompute CIG across tasks. Repeating this procedure produces the Monte Carlo null distribution under directional independence, from which a one-sided $p$-value is obtained by comparing the observed CIG with the simulated values~\citep{hope1968simplified}. As in Level~1, we apply the Benjamini--Hochberg procedure across model pairs and report FDR-adjusted $q$-values.

\noindent
\textbf{Remarks.}
Level~2 complements Level~1 by refining pairwise failure synchronization into directional structure. While Level~1 captures whether two models excessively \emph{fail together}, Level~2 captures whether they excessively fail in the \emph{same way}.

\subsection{Entanglement-Aware Verifier Reweighting}

Beyond auditing behavioral dependence, the proposed metrics can provide a simple signal for adjusting verifier aggregation in multiple LLM judges. Rather than treating all verifier outputs as independent evidence, we examine whether the influence of a verifier can be reduced when it is behaviorally entangled with the target model. We present this reweighting as a downstream use case of the proposed audit.

Let $S$ denote the target model and
$
\mathcal{J}
=
\{J_1,\ldots,J_{M_J}\}
$
the set of verifiers. Each verifier $J_m$ is assigned a competence score $q_m\in(0,1]$, estimated on a calibration set. Since BEI and CIG characterize complementary forms of dependence but operate on different scales, we first normalize their positive components on the calibration set. Let $\widetilde{\mathrm{BEI}}(i,j)$ and $\widetilde{\mathrm{CIG}}(i,j)$ denote the resulting scores. Their combined pairwise entanglement is defined as
\begin{equation}
E(i,j)
=
\lambda
\widetilde{\mathrm{BEI}}(i,j)
+
(1-\lambda)
\widetilde{\mathrm{CIG}}(i,j),
\qquad
\lambda\in[0,1].
\label{eq:combined_entanglement}
\end{equation}

For each verifier, we consider two forms of dependence:
\begin{equation}
R_m
=
\frac{1}{M_J-1}
\sum_{\substack{\ell=1\\\ell\neq m}}^{M_J}
E(J_m,J_\ell),
\qquad
T_m^{(S)}
=
E(J_m,S).
\label{eq:verifier_dependence}
\end{equation}
Here, $R_m$ measures the average redundancy of verifier $J_m$ with the remaining verifier ensemble, while $T_m^{(S)}$ measures its behavioral dependence on the target model.

We then combine verifier competence with penalties for these two forms of dependence:
\begin{equation}
w_m^{(S)}
=
\frac{
q_m^{\kappa}
\exp\!\left(
-\eta_1 R_m
-\eta_2 T_m^{(S)}
\right)
}{
\displaystyle
\sum_{\ell=1}^{M_J}
q_\ell^{\kappa}
\exp\!\left(
-\eta_1 R_\ell
-\eta_2 T_\ell^{(S)}
\right)
}.
\label{eq:entanglement_reweighting}
\end{equation}
The competence term $q_m^{\kappa}$ favors more reliable verifiers, while the two exponential penalties reduce the influence of verifiers that provide potentially redundant evidence. The parameter $\eta_1$ controls the penalty for dependence within the verifier ensemble, and $\eta_2$ controls the penalty for dependence between a verifier and the target model. Setting $\eta_1=\eta_2=0$ recovers competence-based weighting, while additionally setting $\kappa=0$ recovers uniform aggregation. For binary verifier decisions $V_m^{(S)}\in\{0,1\}$, the resulting ensemble score is
\begin{equation}
\widehat{P}^{(S)}
=
\sum_{m=1}^{M_J}
w_m^{(S)}V_m^{(S)},
\label{eq:weighted_verifier_score}
\end{equation}
which replaces the uniform vote share used in majority voting. All normalization procedures and reweighting parameters are determined on the calibration set and fixed before evaluation. This construction tests whether the behavioral dependence identified by BEI and CIG can provide a useful adjustment to verifier aggregation, without claiming to identify or eliminate the underlying source of that dependence.

\section{Experiments}

\subsection{Experimental Setup}

\textbf{Models and Datasets.} We evaluate 18 LLMs spanning the GPT \citep{singh2025openai}, Claude, Qwen~\citep{bai2023qwen}, Llama \citep{grattafiori2024llama}, Gemini \citep{team2023gemini}, and DeepSeek \citep{liu2024deepseek} families to examine behavioral dependence both within and across model families (full list in Appendix B). Three representative models, Llama-3.1-70B-Instruct, GPT-5, and GPT-4o-mini, are further used as judges, covering open-source, high-capability closed-source, and cost-efficient closed-source settings.

Our primary audit uses MMLU-Pro \citep{wang2024mmlu}. We construct two disjoint 1,000-question subsets: biology/psychology (473/527) for estimating pairwise BEI and CIG, and law/business (295/705) for evaluating their association with judge bias. We additionally use MATH-500 \citep{lightman2024let} as a cross-benchmark transfer test, applying the dependence metrics estimated on MMLU-Pro to judge-bias evaluation on non-MCQ mathematical responses. The MATH-500 experiment includes the 14 models that remained available at evaluation time.

\textbf{Implementation Details.}
The experiments consist of two stages: (1) behavioral-dependence estimation and (2) evaluation of dependency-induced judge bias. For answer generation, all models use a unified 5-shot prompt following \citet{wang2024mmlu}, with temperature set to 0 whenever supported. Pairwise BEI and CIG are estimated on the first MMLU-Pro subset using the leave-pair-out null constructions described in Sections~3.1 and~3.2. We compare the proposed metrics against conventional dependence measures, including answer correlation, mutual information, co-failure Jaccard similarity, error-only agreement, and difficulty-adjusted residual correlation.

Judge bias is evaluated on the disjoint second MMLU-Pro subset and the MATH-500 benchmark. Each judge evaluates multiple candidate-model responses within a common prompt, with response order randomly shuffled to reduce ordering effects, and assigns correctness and reasoning-quality scores according to a unified rubric. We additionally conduct a 500-question ablation comparing this joint-prompt
setting with independent single-response judging to assess prompt-induced
stability, with results reported in Appendix~C.3. For verifier reweighting, the 1,000-question evaluation subset is further divided into 500 questions for parameter calibration and 500 held-out questions for final performance reporting.



\begin{figure}[t]
    \centering
    \includegraphics[width=\linewidth,height=0.6\textheight,keepaspectratio]{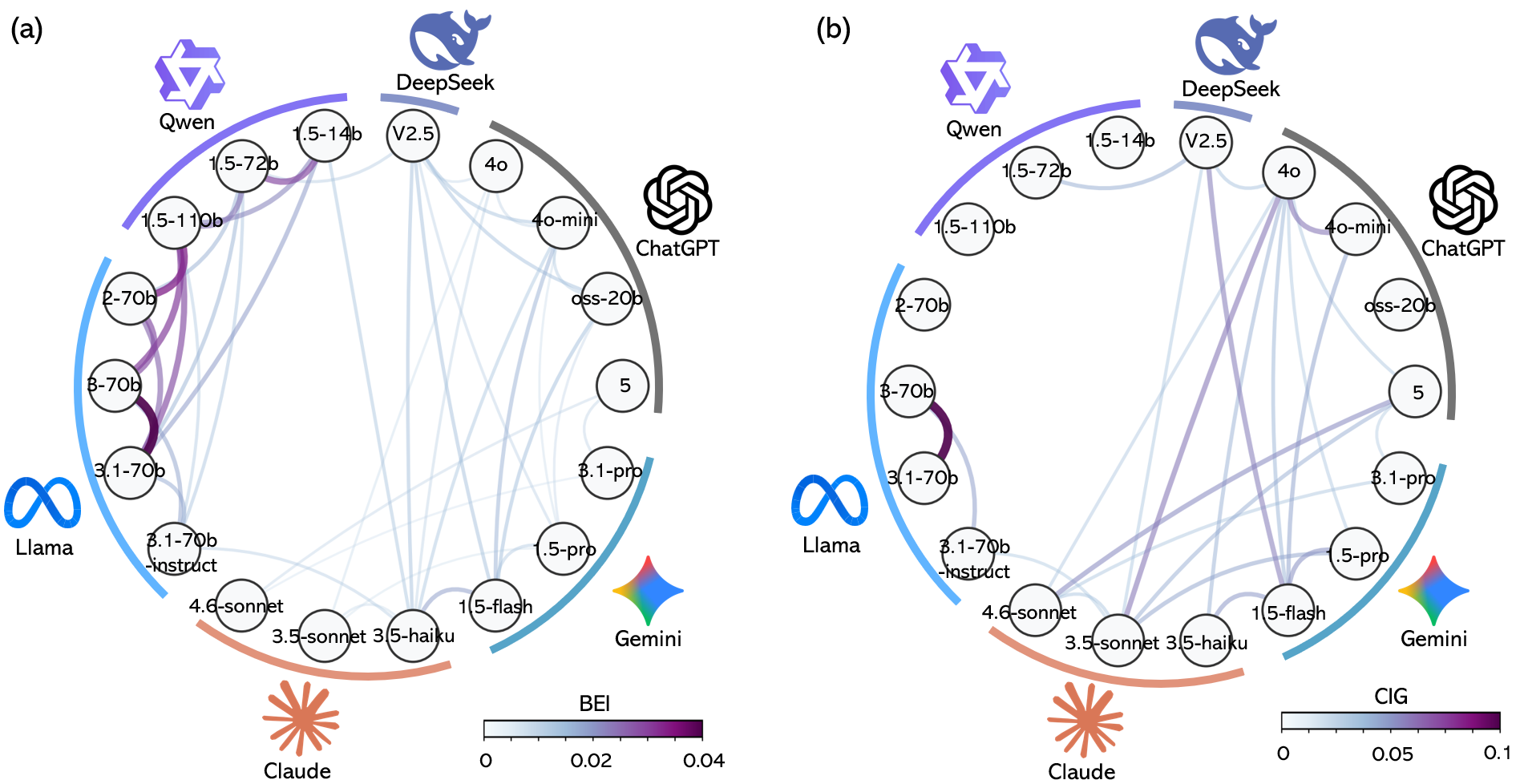}
    \caption{Behavioral entanglement graph of LLMs on MMLU-Pro: (a) co-failure entanglement (BEI); and (b) directional entanglement (CIG). Reported are only statistically significant BEI and CIG values and FDR corrected relationships (q-value $< 0.05$).}
    \label{fig:graph_relation}
\end{figure}

\textbf{Metrics.} We use the deviation in conditional precision ($\Delta\mathrm{Prec}$)
as an operational proxy for over-endorsement bias, measuring how much a
judge's precision for a specific target model deviates from its overall
precision across models:

\begin{equation}
\Delta \mathrm{Prec}(J_j,M_i)
=
\mathbb{P}(Y=1\mid \hat{Y}_j=1)
-
\mathbb{P}(Y=1\mid \hat{Y}_j=1, M=M_i).
\label{eq:delta_precision}
\end{equation}

where $Y=1$ indicates that the target-model response is correct and
$\hat{Y}_j=1$ indicates that judge $J_j$ predicts the response as correct.
A positive $\Delta\mathrm{Prec}$ indicates lower conditional precision for
model $M_i$ relative to the judge's overall precision, consistent with
greater model-specific over-endorsement. We use the Spearman coefficient
to quantify the monotonic association between behavioral dependence and
$\Delta\mathrm{Prec}$.

The proposed BEI and CIG are compared with a diverse set of dependence baselines capturing complementary forms of behavioral association. These include Pearson correlation \citep{pearson1895note}, mutual information \citep{shannon1948}, co-failure Jaccard similarity \citep{jaccard1901}, error-only answer agreement, and difficulty-adjusted residual correlation following the residual-dependence principle of \citet{yen1984}. Detailed definitions of all baseline metrics are provided in Appendix~B.

\subsection{LLM Entanglement and Induced Judge Bias}
\textbf{Entanglement Identification.} We apply our framework to uncover structural relationships across models in the failure space. Using BEI$_w$ and CIG, we construct a behavioral dependency graph (Figure~\ref{fig:graph_relation}), organizing models by shared blind spots and directional error alignment. Based on BEI$_w$, which captures synchronized failures conditioned on task difficulty, we observe strong intra-family entanglement (e.g., within the \emph{LLaMA family}). These findings suggest that behavioral entanglement extends beyond
architectural lineage or model genealogy and may be associated with
shared training signals, alignment strategies, and generation
design choices. Detailed pairwise BEI and CIG statistics, including raw $p$-values and
FDR-adjusted $q$-values, are provided in Appendix~C.2.

While BEI$_w$ provides a coarse view of dependence through correctness and task easiness, CIG offers a more refined perspective by capturing directional collision in errors. Under CIG, the entanglement structure becomes more selective. Consistent patterns are identified by both BEI and CIG, with more statistically significant entanglement pairs observed among closed-weight models and fewer between open- and closed-weight models. In particular, we observe cross-generation entanglement among frontier models (e.g., GPT-5, Claude 4.6, Gemini-3.1-pro), as well as secondary clusters such as Claude 3.5 with GPT-4o, and DeepSeek V2.5 with Gemini 1.5 and Qwen 1.5. These findings suggest that behavioral entanglement extends beyond architectural lineage or model genealogy. These patterns may be associated with shared training signals, alignment
strategies, and generation-era design choices, although the present analysis
does not identify their causal origin.

\begin{table}[t]
\centering
\small
\caption{Metrics and their Spearman correlations with $\Delta{\mathrm{Prec}}$ on MMLU-pro and MATH-500.}
\label{tab:metric_spearman}

\setlength{\tabcolsep}{12pt}
\renewcommand{\arraystretch}{1.08}

\begin{tabular}{lcccc}
\toprule
\multirow{2}{*}{\textbf{Metric}}
& \multicolumn{2}{c}{\textbf{MMLU-pro}}
& \multicolumn{2}{c}{\textbf{MATH-500}} \\
\cmidrule(lr){2-3}
\cmidrule(lr){4-5}
& $\rho$ & $p$-value
& $\rho$ & $p$-value \\
\midrule

\multicolumn{5}{l}{\textit{Co-failure audit}} \\

Corr.
& 0.063 & 0.724
& 0.187 & 0.360 \\

MI
& 0.113 & 0.525
& 0.208 & 0.309 \\

Jaccard
& 0.428 & 0.012$^{*}$
& 0.275 & 0.175 \\

BEI
& 0.508 & 0.002$^{**}$
& 0.441 & 0.024$^{*}$ \\

\midrule

\multicolumn{5}{l}{\textit{Directional failure audit}} \\

Err-agree
& -0.107 & 0.545
& -0.172 & 0.399 \\

Resid. corr.
& 0.151 & 0.394
& -0.057 & 0.746 \\

CIG
& 0.520 & 0.002$^{**}$
& 0.457 & 0.019$^{*}$ \\

\bottomrule

\multicolumn{5}{l}{\footnotesize
$^{*}p<0.05$, $^{**}p<0.01$.}

\end{tabular}
\end{table}

\textbf{Induced Over-endorsement Bias.} We further investigate whether the identified behavioral dependence is associated
with systematic over-endorsement in \textit{LLM-as-a-judge} evaluation. Table~\ref{tab:metric_spearman} compares the proposed metrics with several
conventional dependence measures, including Pearson correlation, mutual
information, co-failure Jaccard similarity, error-only answer agreement, and
difficulty-adjusted residual correlation. On MMLU-Pro, BEI and CIG show the
strongest associations with $\Delta\mathrm{Prec}$, with Spearman coefficients of
$0.508$ and $0.520$, respectively ($p<0.01$). In contrast, Pearson correlation
and mutual information exhibit only weak and non-significant relationships. This contrast suggests that structured failure-based dependence captures aspects of judge--model coupling that are not well represented by conventional aggregate similarity measures, particularly when shared errors contribute to over-endorsement.

Importantly, the same dependency estimates remain predictive when transferred to
MATH-500. BEI and CIG retain significant positive correlations with
$\Delta\mathrm{Prec}$, reaching $\rho=0.441$ and $\rho=0.457$, respectively
($p<0.05$), while most baseline metrics are not significant. These results suggest that the dependency signals identified on MMLU-Pro
are not specific to the original estimation subset and remain informative when transferred to a math benchmark, capturing a transferable structure associated with judge over-endorsement across evaluation settings.

\setlength{\columnsep}{14pt}

\begin{wraptable}{r}{0.52\columnwidth}
  \vspace{-8pt}
  \centering
  \footnotesize
  \caption{Sensitivity of BEI correlations to the difficulty-weighting exponent $w$.}
  \label{tab:bei_sensitivity}

  \setlength{\tabcolsep}{16pt}
  \renewcommand{\arraystretch}{1.02}

  \begin{tabular}{ccc}
    \toprule
    $w$ & MMLU-pro $\rho$ & MATH-500 $\rho$ \\
    \midrule
    0 & 0.289 & 0.032 \\
    1 & 0.508$^{**}$ & 0.441$^{*}$ \\
    2 & 0.490$^{**}$ & 0.275 \\
    3 & 0.504$^{**}$ & 0.231 \\
    4 & 0.465$^{**}$ & 0.236 \\
    5 & 0.363$^{*}$ & 0.180 \\
    \bottomrule
  \end{tabular}

  \vspace{1pt}
  {\scriptsize $^{*}p<0.05$, $^{**}p<0.01$.}

  \vspace{-10pt}
\end{wraptable}

A sensitivity analysis of BEI with respect to the
difficulty-weighting exponent $w$ is conducted (Table~\ref{tab:bei_sensitivity}).
When $w=0$, BEI reduces to the mean co-failure residual. Although this
form is mathematically simple, its empirical association with
$\Delta\mathrm{Prec}$ is weaker, suggesting that unweighted residuals
retain noise from less informative events, particularly outcomes on
difficult tasks. Introducing a positive $w$ places greater emphasis on
co-failures on easier tasks, where synchronized errors provide clearer
evidence of behavioral entanglement. However, overly large $w$
increasingly concentrates the metric on only the easiest tasks and
weakens the association. Across both MMLU-Pro and MATH-500, the results
suggest that moderate weighting, around $w=1$--$2$, provides a
reasonable balance and transfers more consistently across benchmarks.

\subsection{LLM Verifier Ensemble Re-weighting}

Through experiments using three judge models as a verifier ensemble, we evaluate the effectiveness of different aggregation strategies. We compare majority voting with an accuracy-calibrated reweighting baseline
and the proposed entanglement-aware reweighting approach. Results on the
MMLU-Pro test set show that both reweighting strategies outperform majority
voting (Table~3). Notably, our entanglement-aware reweighting achieves the best performance, yielding the highest accuracy and precision.

\begin{table}[t]
\centering
\small
\caption{Verifier aggregation performance on MMLU-Pro.}
\label{tab:verifier_results}

\setlength{\tabcolsep}{6pt}
\renewcommand{\arraystretch}{1.08}

\begin{tabular}{lcccc}
\toprule
\textbf{Aggregation}
& \textbf{Acc}
& \textbf{F1}
& \textbf{Precision}
& \textbf{$\Delta$Acc} \\
\midrule

Majority Vote
& 0.847
& 0.901
& 0.870
& -- \\

Accuracy-based Reweight
& 0.881
& 0.919
& 0.891
& +0.034\; \\

Entangle-based Reweight
& 0.882
& 0.922
& 0.896
& +0.035\; \\

\bottomrule
\end{tabular}
\end{table}

\begin{wrapfigure}{r}{0.55\textwidth}
  \vspace{-10pt}  
  \centering
  \includegraphics[width=0.50\textwidth]{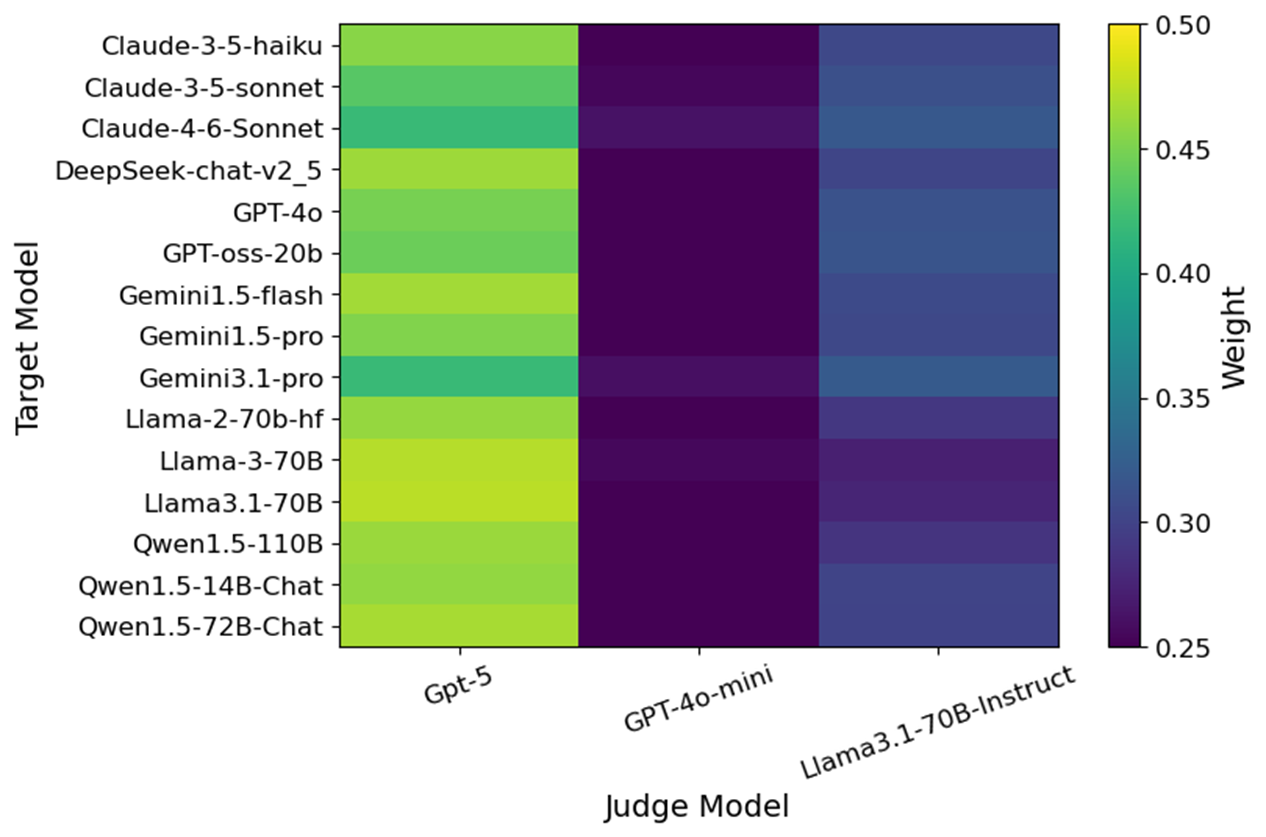}
  \caption{Calibrated verifier ensemble weights.}
  \label{fig:weight}
  \vspace{-14pt}  
\end{wrapfigure}

The calibrated weights (Figure~3) reflect both verifier competence and
the identified dependency structure, with weights adapting across target
models according to verifier redundancy and verifier--target dependence.
This illustrates how the proposed reweighting adjusts verifier influence
according to the estimated dependence structure.

\section{Conclusion}
This work presents a statistical approach for systematically auditing hidden behavioral dependence among LLMs. By shifting the focus from simple output similarity to the structural alignment of the failure manifold, we develop a multi-resolution framework that characterizes dependence through synchronized failures and directional error alignment.

It provides a deeper investigation into the behavioral dependency structure underlying responses generated by different LLMs. It introduces a statistical framework for auditing behavioral entanglement through the failure manifold, using a difficulty-weighted entanglement index (BEI) and a directional error metric (CIG) to detect shared blind spots and aligned erroneous reasoning. Across 18 models from six families, the study reveals widespread intra- and cross-family entanglement, including cross-generation patterns, and demonstrates that stronger dependency is significantly associated with increased judge over-endorsement bias in LLM-as-a-judge settings. It further proposes a practical mitigation strategy, entanglement-aware verifier reweighting, which penalizes dependent models and improves ensemble verification accuracy over majority voting. The paper highlights that Multi-LLM Judge should account for dependency structure, as consensus among entangled models may reflect correlated errors rather than reliable, independent validation signals.

Several limitations remain. The current framework is pairwise, and CIG is most naturally defined for MCQ settings, while MATH-500 evaluates transfer rather than re-estimating directional dependence. Moreover, the proposed audit identifies behavioral association but does not establish its causal origin. Future work should extend the framework to higher-order dependence, broader response formats, and controlled analyses of training lineage, distillation, and temporal model evolution. Finally, our experiments are restricted to English-language benchmarks and represent a snapshot of model behavior at the time of evaluation. Because model APIs and alignment pipelines evolve, estimated dependency structures should not be viewed as permanent model properties. Future work should examine cross-lingual robustness and repeated audits across model versions.

\bibliographystyle{colm2026_conference}
\bibliography{reference}

\appendix
\numberwithin{equation}{section}

\section{Proofs}
\subsection{Proof of Proposition 1}

\noindent
\textit{Proof.}
For a fixed model pair $(i,j)$ and task $t$, let
\[
p_{i,t}=p_i(d_t)=\mathbb{P}(Y_{i,t}=1\mid d_t),
\qquad
p_{j,t}=p_j(d_t)=\mathbb{P}(Y_{j,t}=1\mid d_t).
\]
Under the conditional-independence null $H_0$,
\[
Y_{i,t}\perp\!\!\!\perp Y_{j,t}\mid d_t.
\]
Since $Y_{i,t},Y_{j,t}\in\{0,1\}$, their product indicates a joint
failure. Therefore,
\begin{align}
\mathbb{E}_0[Y_{i,t}Y_{j,t}\mid d_t]
&=
\mathbb{P}_0(Y_{i,t}=1,Y_{j,t}=1\mid d_t) \nonumber\\
&=
\mathbb{P}_0(Y_{i,t}=1\mid d_t)
\mathbb{P}_0(Y_{j,t}=1\mid d_t) \nonumber\\
&=
p_{i,t}p_{j,t}.
\end{align}
Hence, the task-level co-failure residual
\[
R_{ij,t}=Y_{i,t}Y_{j,t}-p_{i,t}p_{j,t}
\]
is conditionally centered:
\[
\mathbb{E}_0[R_{ij,t}\mid d_t]=0.
\]

The BEI weighting term is
\[
a_t^w=(1-d_t)^w,\qquad w\geq0,
\]
which is a deterministic function of $d_t$. Conditional on $d_t$,
it can therefore be taken outside the expectation:
\begin{align}
\mathbb{E}_0[a_t^w R_{ij,t}\mid d_t]
&=
a_t^w\,\mathbb{E}_0[R_{ij,t}\mid d_t]
=0.
\end{align}
By the law of total expectation,
\[
\mathbb{E}_0[a_t^w R_{ij,t}]=0.
\]

Finally, from
\[
\mathrm{BEI}_w(i,j)
=
\frac{1}{T}
\sum_{t=1}^{T}a_t^w R_{ij,t},
\]
linearity of expectation gives
\begin{align}
\mathbb{E}_0[\mathrm{BEI}_w(i,j)]
&=
\frac{1}{T}
\sum_{t=1}^{T}
\mathbb{E}_0[a_t^w R_{ij,t}]
=0.
\end{align}
Thus, weighting by task easiness preserves the null centering of the
co-failure residual. This result does not require independence across
tasks, since it follows task-wise from conditional independence and
linearity of expectation.

\section{Experimental Details}
\subsection{Selected models}

We select 18 models from the ChatGPT, Claude, Qwen, Llama, Gemini, and DeepSeek families to examine potential entanglement both within and across model families. Specifically, the selected models are GPT-5 \citep{singh2025openai}, GPT-4o, GPT-4o-mini \citep{hurst2024gpt}, GPT-oss-20B \citep{agarwal2025gpt}, Claude 4.6 Sonnet, Claude-3.5-Sonnet, Claude-3.5-Haiku, Gemini-3.1-Pro \citep{team2023gemini}, Gemini-1.5-Pro~\citep{team2024gemini}, Gemini-1.5-Flash~\citep{team2024gemini}, Llama-3.1-70B~\citep{grattafiori2024llama}, Llama-3.1-70B(-Instruct)~\citep{grattafiori2024llama}, Llama-3-70B  \citep{grattafiori2024llama}, Llama-2-70b~\citep{touvron2023llama}, Qwen1.5-110B~\citep{bai2023qwen}, Qwen1.5-72B Chat, Qwen1.5-14B Chat~\citep{bai2023qwen}, and DeepSeek-Chat-v2.5~\citep{liu2024deepseek}, where “Instruct” denotes instruction-tuned variants.

\subsection{Baseline Dependence Metrics}
\label{app:baseline_metrics}

We compare BEI and CIG against five conventional dependence measures that capture complementary aspects of model behavior. Let
$Y_{m,t}\in\{0,1\}$ denote the failure indicator of model $m$ on task $t$, and let
\[
\mathcal{F}_m=\{t:Y_{m,t}=1\}
\]
denote its set of failed tasks.

Pearson correlation. We measure the overall linear association between the binary failure patterns of models $i$ and $j$ using the Pearson correlation coefficient \citep{pearson1895note}:
\begin{equation}
\mathrm{Corr}(i,j)
=
\frac{
\sum_{t=1}^{T}
(Y_{i,t}-\bar{Y}_i)
(Y_{j,t}-\bar{Y}_j)
}{
\sqrt{
\sum_{t=1}^{T}(Y_{i,t}-\bar{Y}_i)^2
\sum_{t=1}^{T}(Y_{j,t}-\bar{Y}_j)^2
}
},
\label{eq:baseline_corr}
\end{equation}
where
$\bar{Y}_m=T^{-1}\sum_{t=1}^{T}Y_{m,t}$.

Mutual information.
Mutual information measures general statistical dependence between the two binary failure indicators \citep{shannon1948}. Let
\[
\hat{P}_{ab}
=
\frac{1}{T}
\sum_{t=1}^{T}
\mathbb{I}(Y_{i,t}=a,Y_{j,t}=b),
\qquad a,b\in\{0,1\},
\]
with marginal probabilities
$\hat{P}_{a\cdot}=\sum_b\hat{P}_{ab}$ and
$\hat{P}_{\cdot b}=\sum_a\hat{P}_{ab}$.
We compute
\begin{equation}
\mathrm{MI}(i,j)
=
\sum_{a,b\in\{0,1\}}
\hat{P}_{ab}
\log
\frac{\hat{P}_{ab}}
{\hat{P}_{a\cdot}\hat{P}_{\cdot b}}.
\label{eq:baseline_mi}
\end{equation}

Co-failure Jaccard similarity.
To measure overlap specifically between model failure sets, we use the Jaccard similarity coefficient \citep{jaccard1901}:
\begin{equation}
\mathrm{Jac}(i,j)
=
\frac{
|\mathcal{F}_i\cap\mathcal{F}_j|
}{
|\mathcal{F}_i\cup\mathcal{F}_j|
}
=
\frac{
\sum_{t=1}^{T}Y_{i,t}Y_{j,t}
}{
\sum_{t=1}^{T}
\mathbb{I}(Y_{i,t}+Y_{j,t}>0)
}.
\label{eq:baseline_jaccard}
\end{equation}

Error-only answer agreement.
For MCQ tasks, let
$\mathcal{T}_{ij}^{\mathrm{cf}}
=
\{t:Y_{i,t}=Y_{j,t}=1\}$
denote the set of co-failure tasks, and let $S_{m,t}$ denote the distractor selected by model $m$.
The error-only agreement rate is
\begin{equation}
\mathrm{ErrAgree}(i,j)
=
\frac{1}{
|\mathcal{T}_{ij}^{\mathrm{cf}}|
}
\sum_{t\in\mathcal{T}_{ij}^{\mathrm{cf}}}
\mathbb{I}(S_{i,t}=S_{j,t}).
\label{eq:baseline_error_agreement}
\end{equation}
Unlike CIG, this baseline treats all same-distractor collisions equally and does not account for task-specific distractor attractiveness.

Difficulty-adjusted residual correlation.
Finally, we consider an IRT-style residual-correlation baseline \citep{yen1984}. Using the fitted difficulty-response probability
$\hat{p}_{m,t}=\hat{p}_m(d_t)$, define
\begin{equation}
e_{m,t}
=
Y_{m,t}-\hat{p}_{m,t}.
\label{eq:baseline_residual}
\end{equation}
The residual dependence between models $i$ and $j$ is then measured by
\begin{equation}
\mathrm{ResCorr}(i,j)
=
\mathrm{Corr}
\left(
\{e_{i,t}\}_{t=1}^{T},
\{e_{j,t}\}_{t=1}^{T}
\right).
\label{eq:baseline_rescorr}
\end{equation}
This baseline removes first-order variation associated with task difficulty before measuring pairwise association.

\subsection{Judge and Verifier Prompts}

Each model is required to evaluate the performance of all other models under a unified cross-evaluation protocol. To mitigate ordering effects and prevent positional bias, the sequence of evaluated model responses is randomly shuffled for each query. All responses are presented within a single prompt to ensure consistent evaluation context across models. The verifier is instructed to independently assess each response based on two criteria: (i) correctness of the selected answer and (ii) quality of the reasoning process, following strict rubrics to enforce consistency and comparability. The evaluation is designed to be objective and model-agnostic, avoiding reliance on stylistic preferences or verbosity. By standardizing both input formatting and evaluation instructions, this setup minimizes prompt-induced variability and ensures that observed differences in judgments reflect model behavior rather than evaluation artifacts.

\begin{tcolorbox}[
  colback = blue!10!white,
  colframe = blue!60!black,
  title   = {\strut Prompt for Cross Evaluation},
  width   = \linewidth, top = 2pt, bottom = 2pt]
\small

\textbf{Evaluation Criterion}

You are a fair, consistent, and rigorous evaluator of multiple LLM responses.  
Your task is to evaluate the answers provided by different models to the same multiple-choice question.

For each model, you must assess:
\begin{itemize}
    \item \textbf{Correctness} (\textcolor{blue}{0} or \textcolor{blue}{1}) of the \textcolor{blue}{Selected Option}
    \item \textbf{Reasoning Quality} (\textcolor{blue}{0--5}) of the \textcolor{blue}{Reasoning Process}
\end{itemize}

Follow the provided rubrics strictly. Your evaluation must be objective, consistent, and comparable across models.\\[6pt]
\textbf{Input}

Question: \textcolor{blue}{\{QA.\textit{\textless key\textgreater}.question\}}
Options: \textcolor{blue}{\{QA.\textit{\textless key\textgreater}.options\}}\\[6pt]
Model1 selects: \textcolor{blue}{\{QA.\textit{\textless key\textgreater}.answers.LLM1.selected\_option\}}\\
Model1 reasoning: \textcolor{blue}{\{QA.\textit{\textless key\textgreater}.answers.LLM1.reasoning\}}\\[6pt]
Model2 selects: \textcolor{blue}{\{QA.\textit{\textless key\textgreater}.answers.LLM2.selected\_option\}}\\
Model2 reasoning: \textcolor{blue}{\{QA.\textit{\textless key\textgreater}.answers.LLM2.reasoning\}}\\[6pt]
... (extend for all models)\\[6pt]
\textbf{Evaluation Rubrics}

Correctness rubric: \textcolor{blue}{\{Cross Evaluation Rubric: Correctness\}}\\
Reasoning Quality rubric: \textcolor{blue}{\{Cross Evaluation Rubric: Reasoning Quality\}}\\[6pt]
\textbf{Output Format (STRICT)}

Return a valid \textbf{JSON object} with the following structure:
{\color{blue}
\begin{verbatim}
{
  "correct_answer": "<final correct option>",
  "evaluation_summary": "<comparison of model performance>",
  "evaluation": {
    "model1": {
      "correctness": <0 or 1>,
      "reasoning_quality": <0-5>
    },
    "model2": {
      "correctness": <0 or 1>,
      "reasoning_quality": <0-5>
    },
    ...
  }
}


\end{verbatim}
}

\end{tcolorbox}

\begin{tcolorbox}[
  colback = blue!10!white,
  colframe = blue!60!black,
  title   = {\strut Cross Evaluation Rubric: Correctness},
  width   = \linewidth, top = 2pt, bottom = 2pt]
\small

\textbf{Evaluation Criteria}

Given the question, candidate options, and the \textcolor{blue}{Selected Option} produced by each model, evaluate whether the model's final answer is correct. The correctness judgment should focus \textbf{only} on the choice of option, rather than the explanation.

\begin{itemize}
    \item \textcolor{blue}{\textbf{1}} $\rightarrow$ Correct
    \item \textcolor{blue}{\textbf{0}} $\rightarrow$ Incorrect or missing
\end{itemize}

\textbf{Rules}

\begin{itemize}
    \item Evaluate correctness based on the model's selected option.
    \item If the model explicitly states a single option that matches the correct answer, assign \textbf{1}.
    \item If the model explicitly states a single option that does not match the correct answer, assign \textbf{0}.
    \item If the response does not contain a clear final answer, assign \textbf{0}.
\end{itemize}

\end{tcolorbox}

\begin{tcolorbox}[
  colback = blue!10!white,
  colframe = blue!60!black,
  title   = {\strut Cross Evaluation Rubric: Reasoning Quality},
  width   = \linewidth, top = 2pt, bottom = 2pt]
\small

\textbf{Evaluation Criteria}

Given the model's full response, evaluate the quality of its \textcolor{blue}{Reasoning Process} based on the logical validity, coherence, relevance, and completeness of the explanation that supports the final answer. The reasoning score should reflect \textbf{how well the model reasons}, rather than simply on whether the final answer is correct.

Assign a reasoning quality score on a 0--5 scale:

\begin{itemize}
    \item \textcolor{blue}{\textbf{5}} $\rightarrow$ Fully correct, clear, logically sound, and well-supported reasoning with no hallucinations or unnecessary steps
    \item \textcolor{blue}{\textbf{4}} $\rightarrow$ Mostly correct and coherent reasoning with only minor issues, omissions, or inefficiencies
    \item \textcolor{blue}{\textbf{3}} $\rightarrow$ Partially correct reasoning with some weak, unclear, incomplete, or loosely justified steps
    \item \textcolor{blue}{\textbf{2}} $\rightarrow$ Reasoning contains major flaws, missing key logical steps, or substantial unsupported claims
    \item \textcolor{blue}{\textbf{1}} $\rightarrow$ Mostly incorrect, confused, or poorly connected reasoning with minimal valid support
    \item \textcolor{blue}{\textbf{0}} $\rightarrow$ No meaningful reasoning, completely nonsensical content, or irrelevant response
\end{itemize}

\textbf{Rules}

\begin{itemize}
    \item A correct final answer with poor reasoning can still receive a low reasoning score.
    \item An incorrect final answer may still receive a moderate reasoning score if the reasoning is largely logical but contains a critical mistake.
    \item Focus on whether each step follows logically from the previous one and whether the reasoning is relevant to the question.
    \item Do \textbf{not} reward verbosity, repetition, or persuasive language. Penalize hallucinated facts, logical gaps, invalid assumptions, or unsupported claims.
    \item Use a \textbf{strict but fair} standard: high scores should only be given when the reasoning is both logically sound and clearly articulated.
\end{itemize}

\end{tcolorbox}

\subsection{Reweighting Calibration and Hyperparameters}

For verifier reweighting, we use a separated calibration and evaluation procedure to avoid tuning on the reported test results. The 1,000-question MMLU-Pro evaluation subset is divided into 500 questions for calibration and 500 held-out questions for final evaluation. On the calibration split, we first estimate each verifier's competence score $q_m$ and normalize the positive BEI and CIG values used to construct the pairwise entanglement score. The reweighting parameters $\lambda$, $\kappa$, $\eta_1$, and $\eta_2$ are then calibrated by minimizing binary cross-entropy between the weighted verifier ensemble prediction and the ground-truth correctness label of each target-model response. Here, $\lambda$ controls the relative contribution of BEI and CIG, $\kappa$ controls the influence of verifier competence, and $\eta_1$ and $\eta_2$ penalize verifier--verifier redundancy and verifier--target dependence, respectively. After calibration, all normalization statistics and hyperparameters are fixed and applied unchanged to the held-out evaluation split.

\section{Supplementary Results}

\subsection{Difficulty-response Function Calibration in BEI Calculation}

To estimate the difficulty response function 
\( p_m(d_t) = \mathbb{P}(Y_{m,t}=1 \mid d_t) \), 
we fit a logistic regression model for each model \(m\), using task difficulty \(d_t\) as the input and the binary failure indicator \(Y_{m,t}\) as the target. This provides a calibrated estimate of each model’s expected failure probability conditioned on task difficulty, which is used to compute the residuals.

\begin{table}[ht]
\centering
\small
\caption*{Table C.1: Calibration diagnostics for the difficulty-response models used in BEI.}
\label{tab:bei_calibration}

\setlength{\tabcolsep}{10pt}
\renewcommand{\arraystretch}{1.08}

\begin{tabular}{lccc}
\toprule
\textbf{Metric} & \textbf{Mean} & \textbf{Std.} & \textbf{Range} \\
\midrule
AUC   & 0.865 & 0.050 & [0.752, 0.934] \\
ECE   & 0.041 & 0.012 & [0.022, 0.077] \\
Brier & 0.130 & 0.039 & [0.065, 0.192] \\
\bottomrule
\end{tabular}
\end{table}

\begin{figure}[ht]
    \centering\includegraphics[width=0.8\linewidth,height=0.6\textheight,keepaspectratio]{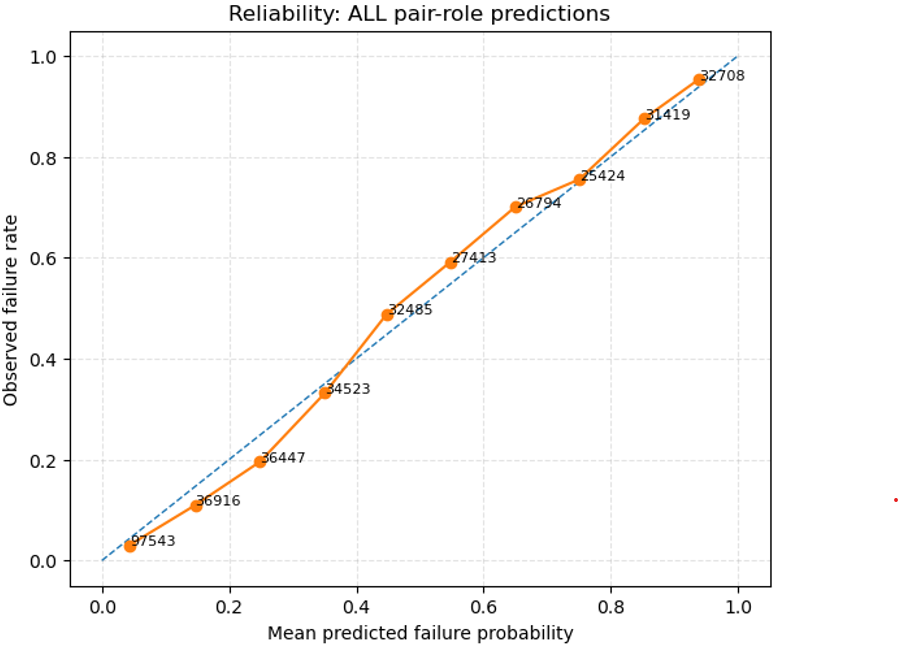}
    \caption*{Figure C.1: Reliability curve of the difficulty-response models used for BEI calibration.}
    \label{fig:Entanglement Hierarchy}
\end{figure}

We evaluate the difficulty-response calibration using AUC, ECE, and
Brier score, together with the reliability curves shown in
Figure~C.1. As summarized in Table~C.1, the fitted models achieve a
mean AUC of 0.865, a low mean ECE of 0.041, and a mean Brier score of
0.130. Together with the reliability curves, these results indicate
that task difficulty provides strong discrimination of model failures
while the estimated failure probabilities remain reasonably well
calibrated for the subsequent BEI calculation.

\subsection{Pairwise Statistical Inference and FDR Correction}

To complement the aggregate dependency analysis, we report the strongest
pairwise BEI and CIG relationships together with their raw $p$-values and
Benjamini--Hochberg FDR-adjusted $q$-values. This allows the magnitude of
behavioral dependence to be distinguished from its statistical reliability
under multiple pairwise comparisons.

Table~C.2 shows the largest BEI values. The strongest relationships are
concentrated primarily among closely related model families, including
Llama--Llama and Qwen--Qwen pairs, while several cross-family relationships
are also observed. Importantly, the leading BEI relationships remain
statistically significant after FDR correction, supporting the robustness of
the detected co-failure dependencies.

\begin{table}[ht]
\centering
\small
\setlength{\tabcolsep}{5pt}
\renewcommand{\arraystretch}{1.1}
\caption*{Table C.2: Top 10 pairwise BEI values with raw $p$-values and
FDR-adjusted $q$-values.}
\begin{tabular}{l l c c c}
\toprule
\textbf{Model 1} & \textbf{Model 2} & \textbf{BEI} 
& \textbf{$p$-value} & \textbf{$q$-value} \\
\midrule

Llama-3-70B
& Llama-3\_1-70B
& 0.0525
& 5.00E-05
& 9.50E-04 \\

Llama-2-70b-hf
& Qwen1.5-110B
& 0.0398
& 5.00E-05
& 9.50E-04 \\

Llama-3-70B
& Qwen1.5-110B
& 0.0352
& 5.00E-05
& 9.50E-04 \\

Qwen1.5-14B-Chat
& Qwen1.5-72B-Chat
& 0.0333
& 5.00E-05
& 9.50E-04 \\

Llama-2-70b-hf
& Llama-3-70B
& 0.0326
& 5.00E-05
& 9.50E-04 \\

Llama-3\_1-70B
& Qwen1.5-110B
& 0.0325
& 5.00E-05
& 9.50E-04 \\

Llama-2-70b-hf
& Llama-3\_1-70B
& 0.0290
& 1.50E-04
& 1.97E-03 \\

Qwen1.5-110B
& Qwen1.5-72B-Chat
& 0.0266
& 2.50E-04
& 2.38E-03 \\

Qwen1.5-110B
& Qwen1.5-14B-Chat
& 0.0245
& 8.50E-04
& 6.61E-03 \\

Llama-3\_1-70B
& Qwen1.5-14B-Chat
& 0.0211
& 2.15E-03
& 1.31E-02 \\

\bottomrule
\end{tabular}

\vspace{4pt}

\label{tab:all_gbei_pairs}
\end{table}

Table~C.3 reports the corresponding results for CIG. Compared with BEI, the
largest CIG values reveal a more selective directional dependency structure,
including both intra-family relationships, such as Llama--Llama and
GPT--GPT, and cross-family relationships involving Claude, Gemini, and
DeepSeek models. The leading CIG relationships likewise retain significance
after FDR adjustment.

\begin{table}[ht]
\centering
\small
\setlength{\tabcolsep}{5pt}
\renewcommand{\arraystretch}{1.1}
\caption*{Table C.3: Top pairwise CIG values with raw $p$-values and
FDR-adjusted $q$-values.}
\begin{tabular}{l l c c c}
\toprule
\textbf{Model 1} & \textbf{Model 2} & \textbf{CIG} 
& \textbf{$p$-value} & \textbf{$q$-value} \\
\midrule

Llama-3-70B
& Llama-3\_1-70B
& 0.1332
& 5.00E-05
& 1.09E-03 \\

Claude-3-5-sonnet-20241022
& GPT-4o-2024-08-06
& 0.0525
& 5.00E-05
& 1.09E-03 \\

GPT-4o-2024-08-06
& GPT-4o-mini
& 0.0523
& 5.00E-05
& 1.09E-03 \\

DeepSeek-chat-v2\_5
& Gemini-1.5-flash-002
& 0.0502
& 5.00E-05
& 1.09E-03 \\

Claude-4-6-Sonnet
& Gpt-5
& 0.0471
& 5.00E-05
& 1.09E-03 \\

Gemini-1.5-flash-002
& Gemini-1.5-pro-002
& 0.0454
& 5.00E-05
& 1.09E-03 \\

Claude-3-5-haiku-20241022
& Gemini-1.5-flash-002
& 0.0429
& 2.00E-04
& 3.40E-03 \\

Llama-3-70B
& Llama-3\_1-70B-Instruct
& 0.0414
& 2.50E-04
& 3.83E-03 \\

Gemini-1.5-flash-002
& GPT-4o-mini
& 0.0393
& 8.50E-04
& 8.61E-03 \\

\bottomrule
\end{tabular}

\vspace{4pt}

\label{tab:all_cig_pairs}
\end{table}

\subsection{Joint-Prompt vs.\ Independent-Prompt Ablation}

We compare the joint-prompt evaluation with independent single-response
judging on a 500-question subset. Let $V_1$ and $V_2$ denote the joint-prompt and independent-prompt
settings, respectively. As shown in Table~C.4, the paired
$t$-test finds no significant shift in $\Delta\mathrm{Prec}$ ($p=0.421$),
while the rank-equivalence test supports consistent target-model rankings
across the two prompting settings ($p=7.6\times10^{-5}$). Overall, the
aggregate judge-bias pattern remains stable under independent prompting.

\begin{table}[ht]
\centering
\small
\setlength{\tabcolsep}{4.5pt}
\renewcommand{\arraystretch}{1.08}

\caption*{Table C.4: Joint-prompt versus independent-prompt evaluation ablation.}

\begin{tabular}{l l l c l}
\toprule
\textbf{Target}
& \textbf{Test}
& \textbf{Null hypothesis}
& \textbf{$p$-value}
& \textbf{Interpretation} \\
\midrule

$\Delta\mathrm{Prec}$
& Paired $t$-test
& $\mathbb{E}[p_t(V_2)-p_t(V_1)]=0$
& 0.421
& No mean shift \\

Ranking
& Equivalence test
& $\rho \leq 1-\epsilon,\ \epsilon=0.05$
& $7.6\times10^{-5}$
& Ranking consistent \\

\bottomrule
\end{tabular}

\label{tab:prompt_ablation}
\end{table}

\end{document}